\documentclass[sigplan,10pt]{acmart}
\usepackage{times}
\usepackage[T1]{fontenc}
\usepackage[scaled]{beramono}
\renewcommand\footnotetextcopyrightpermission[1]{}
\usepackage{packages}
\newcommand{\para}[1]{\smallskip \noindent \textbf{{#1. } }}
\newcommand{\secref}[1]{(\S\ref{#1})}
\newcommand{\X}{$\times$ \xspace}
\newcommand{\eg}{e.g., \xspace}

\newcommand{\sys}{Alto\xspace}

\newcommand{\cc}[1]{\texttt{#1}}

\setcopyright{none}
\renewcommand\footnotetextcopyrightpermission[1]{} 

\title{\LARGE \sys: Orchestrating Distributed Compound AI Systems with Nested
Ancestry}
\author{
    \normalsize
    Deepti Raghavan$^1$,
    Keshav Santhanam$^2$,
Muhammad Shahir Rahman$^3$,
Nayani Modugula$^1$,
Luis Gaspar Schroeder$^4$,
Maximilien Cura$^3$,
Houjun Liu$^3$,
Pratiksha Thaker$^5$,
Philip Levis$^3$,
Matei Zaharia$^4$\\
Brown University$^1$,\xspace Nvidia$^2$,\xspace Stanford University$^3$,\xspace UC
Berkeley$^4$,\xspace Carnegie Mellon University$^5$
}

\begin{abstract}
  Compound AI applications chain together subcomponents such as
  generative language models, document retrievers, and embedding models.
  Applying traditional systems optimizations such as parallelism and pipelining
  in compound AI systems is difficult
  because each component has different constraints in terms of the
  granularity and type of data that it ingests.
  New data is often generated during intermediate computations,
  and text streams may be split into smaller, independent fragments (such as documents to sentences)
  which may then be re-aggregated at later parts of the computation.
  Due to this complexity, existing systems to serve compound AI queries
  do not fully take advantage of parallelism and pipelining opportunities.
  
  We present \sys, a framework that automatically optimizes 
  execution of compound AI queries through streaming and parallelism.
  \sys introduces a new abstraction called \emph{nested ancestry},
  a metadata hierarchy that allows the system to correctly track
  partial outputs and aggregate data across the heterogeneous
  constraints of the components of compound AI applications.
  This metadata is automatically inferred from the programming model,
  allowing developers to express complex dataflow patterns without needing
  to reason manually about the details of routing and aggregation.
  Implementations of four applications in \sys outperform or match
  implementations in LangGraph, a popular existing AI programming framework.
  \sys implementations match or improve latency by between 10-30\%.
\end{abstract}

\begin{document}
\maketitle
\pagestyle{plain}


\section{Introduction}
\label{section:introduction}
\begin{figure}[t]
\centering
\includegraphics[width=\columnwidth]{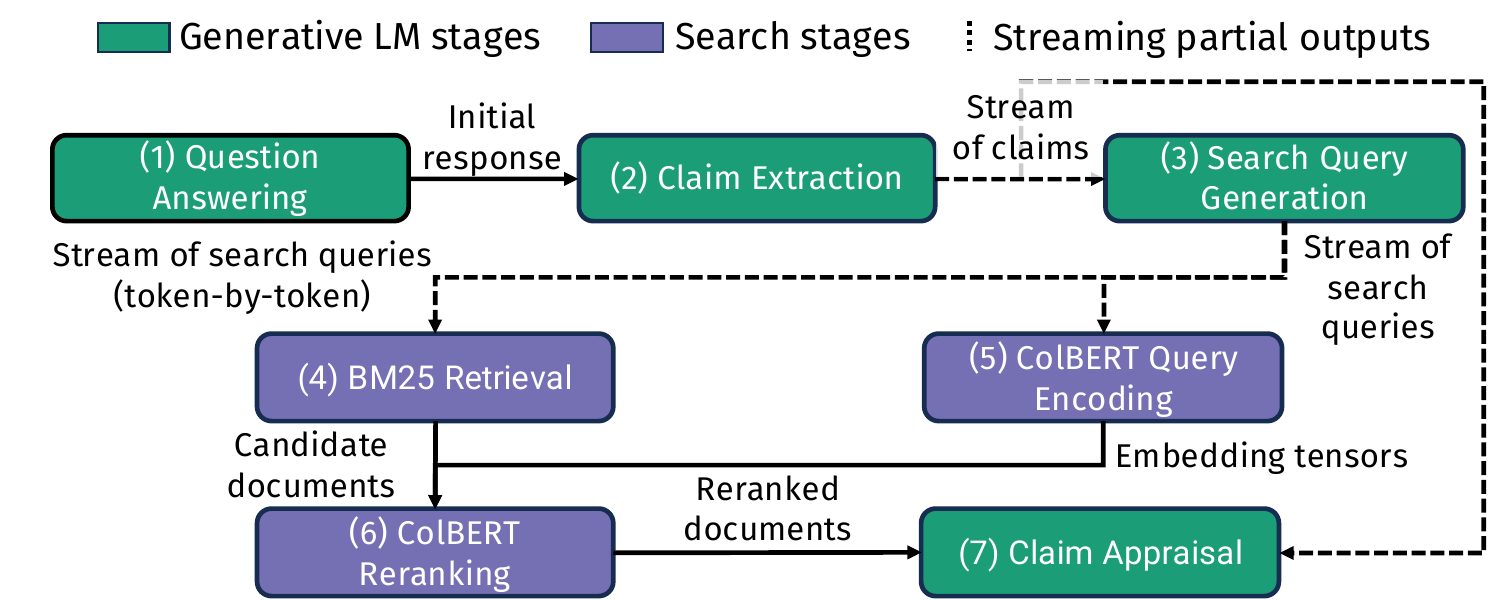}
        \caption{FacTool~\cite{chern2023factool}-inspired pipeline for
          verifying chatbot claims.  For each question
          (stage 1), claims are extracted from the response
          (stage 2) and search queries are generated for each claim
          (stage 3). The search queries retrieve relevant documents
          from a knowledge corpus using BM25 (stage 4) and
          reranked by ColBERT (stages 5 and 6). Finally the claim's accuracy
	  is verified using the retrieved evidence (stage 7).}
\label{fig:factool_diagram}
\end{figure}

AI developers are increasingly using generative language models (LMs)
to program compound AI
systems~\cite{zaharia2024compound,openaiagents,anthropicagents},
which chain together LMs and other services into complex, distributed
applications.
Examples include retrieval-augmented generation
(RAG)~\cite{lewis2020rag, izacard2021leveraging, huang2023raven,
  ram2023context}, structured prompting~\cite{wang2022self,
  yao2023tree, besta2023graph}, multi-hop question
answering~\cite{yao2022react, khattab2023dspy},
agents~\cite{patil2023gorilla, liu2023bolaa, ruan2023tptu, wu2023autogen},
chatbot verification~\cite{chern2023factool, dhuliawala2023chain,
  chen2023complex, semnani2023wikichat}, 
and data querying AIs~\cite{li2023can, sun2023sql}.
For instance, many recent systems use retrieval for factual long-form
generation~\cite{chern2023factool,semnani2023wikichat, storm}.
As a concrete example, FacTool~\cite{chern2023factool} ``fact-checks'' the output
of a generative language model to assess which claims are grounded in a document corpus.
FacTool, shown in Figure~\ref{fig:factool_diagram}, makes this assessment by invoking a pipeline
of LMs and other services. First, it uses
  a LM call to extract claims
  from the output. It uses a second LM call to transform these
claims into search queries, which it uses to request documents
and then rank their contents. Finally, it invokes a third LM call
  to judge whether the documents support the claim.

Generative language models differ from other network services in two
important ways. First, they
produce natural language, which has structure, in the form of words,
sentences, or paragraphs, as well as semantic meaning to
ordering.
Second, instead of always producing
complete outputs, autoregressive models incrementally produce fragments of output, 
emitting a single output token (a word or partial word) in each iteration.
To reduce latency, a subsequent service can process partial text outputs as the
language model generates text.
For example, within FacTool, a document search module can look up the embedding 
for each search word as a language model generates complete search queries.

Outputting and processing partial outputs is an opportunity to stream and
pipeline application execution to reduce latency.
One stage can begin processing output fragments as they are generated, rather
than waiting for the previous stage to complete and process its input in
entirety.
Existing frameworks for building and serving compound AI systems largely miss
this opportunity and rely on blocking input-output relationships
inherited from traditional distributed systems. 

Streaming partial outputs has two major challenges. 
The first is managing the complex and changing granularity of
data passed through the system. Services ingest and produce data at a wide range of
granularities, and the unit of incremental processing is often
distinct from the unit of a complete request.
Returning to the document
search above, while it can incrementally process words, it cannot 
produce its first result until it receives the last word. 
While some services generate and process text 
token-by-token, others require larger units.
An external search engine, for
example, may require an entire query as input and produce whole documents.

The second challenge is how data dependencies and granularity 
constrain parallelization and aggregation of parallel results.  
Some stages produce a series of partial outputs that can
be parallelized over multiple instances of the next step in the pipeline, while
others produce an output whose elements must be serially processed by a single next
stage instance. In FacTool, for example, the search words for a given query must
all be sent to the same document search instance, but different queries from
the same request can be parallelized over multiple document search instances.
Parallelized streams of text processing
must also be joined together, producing aggregate results.
As a result, executing a compound AI application
requires carefully coordinating data as services split, stream, parallelize,
and join data over many different machines.

This paper proposes a data streaming model and programming
interface that allows developers to easily
write and compose complex compound AI systems, which the system automatically
pipelines, streams, and parallelizes across a distributed set of workers. The
core system abstraction that makes this possible is a metadata tag we call {\em nested ancestry}.
Nested ancestry tracks the provenance of data through a computation graph, allowing
the system to manage parallelism and aggregate streaming computations as requests complete,
rather than blocking on the end of a stage.


Because tracking and maintaining nested ancestry can be laborious and
error-prone, we present a programming model that couples it with
program structure. The programming model emphasizes streams as a
basic data structure for processing partial outputs, integrating them
with incremental processing functions similar to iterators. The
programming model introduces the concept of
a \emph{data hierarchy}, representing the relationships between
elements both with their larger data structures and processing
parallelism. The {\tt pmap} (parallel map) operator translates
this data hierarchy into nested ancestry. It takes an ordered stream of text units
and runs function on each unit in parallel.  Each
invocation of {\tt pmap} adds a level of ancestry: the hierarchy
of ancestry tags on any unit of data represents the tree of {\tt pmap}
calls it belongs to as well as its ordering within each of those calls.


We have implemented \sys, a programming model and
orchestration system for compound AI systems that uses nested ancestry
to correctly stream, parallelize, route, split, merge, 
and process fragments of data streams across many layers of system composition.
Programmers write compound AI stages as functions, which can have streams as
inputs or outputs, and invoke LMs or other data
processing elements.
\sys uses nested ancestry to stream and distribute processing over independent
replicas of each stage, automatically reducing latency and increasing
throughput by performing \emph{intra-query pipelining}.

\begin{sloppypar}
We evaluate \sys with four compound AI applications from
the NLP literature:
FacTool~\cite{chern2023factool},
WikiChat~\cite{semnani2023wikichat},
STORM~\cite{storm}, and
Tree Of Thought~\cite{treeofthought}.
\sys implementations match or improve performance of hand-optimized implementations of these applications written in
LangGraph~\cite{chase2022langgraph}.
At best, for FacTool, \sys provides 25\% lower median latency at low load, and 2.3\X higher throughput at the median latency SLO of 6 seconds.
At worst, as STORM does not benefit from pipelining, \sys maintains
the same latency.
\end{sloppypar}

This paper makes three contributions.
First, it proposes nested ancestry
as an abstraction for orchestrating and tracking the execution of complex
compound AI pipelines.
Second, it shows how nested ancestry can be tightly
coupled with a programming model, such that ancestry can be inferred from
program structure.
Finally, it demonstrates that nested ancestry's information
allows an orchestration system to automatically and correctly distribute and
stream complex applications, providing latency and throughput benefits.

\section{Compound AI Applications}
\label{sec:appsandmotivation}
This section outlines the properties of compound AI applications and derives
three design requirements: data hierarchy, parallelism and streaming, and
composability.
Existing frameworks like LangGraph~\cite{chase2022langgraph},DSPy~\cite{khattab2023dspy}
Autogen~\cite{wu2023autogen}, or Llamaindex~\cite{liu2022llamaindex} support
these applications 
but lack end-to-end optimizations across LM and non-LM components.

\subsection{Properties of Compound AI Applications}
\label{sec:appsandmotivation:properties}

Compound AI applications exhibit properties that
distinguish them from prior distributed systems workloads
such as single-model ML serving or batch data analytics.
The generative language models central to many compound AI applications
produce streams of natural language text that can be processed at multiple
granularities. For example, a text stream can be delimited at the boundary
of paragraphs, sentences, new lines, or individual words. Decomposing the
text streams into these partial outputs helps pipeline computation.

Moreover, individual partial outputs can be processed
in parallel by fanning them out across downstream stage instances --
which also enables fine-grained load balancing.
The fan-out of partial outputs in a streaming, parallelized compound AI
application induces an inherent hierarchy across intermediate data objects.
A pipeline may not just invoke one stage on a partial output; it may invoke
an entire subgraph, on each partial output.
These subgraphs can also produce streams which fan-out into a nested subgraph.

\begin{figure}[t!]
    \centering
    \begin{subfigure}[t]{0.6\columnwidth}
        \includegraphics[width=\columnwidth]{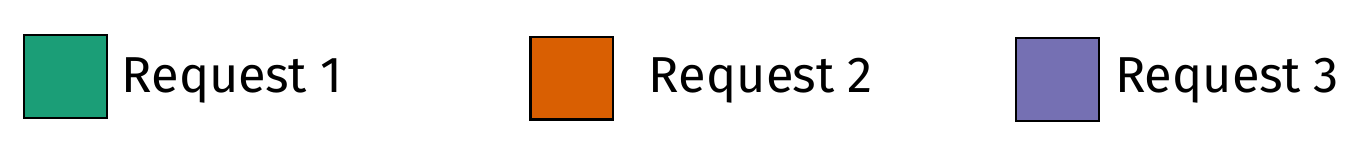}
    \end{subfigure}
    \begin{subfigure}[t]{\columnwidth}
        \includegraphics[width=\columnwidth]{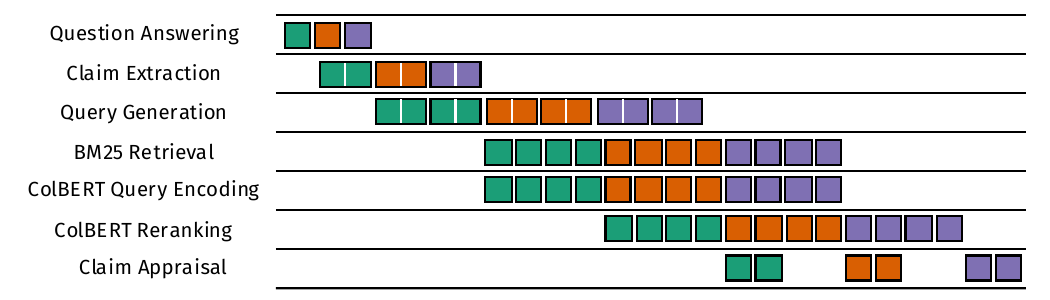}
        \caption{FacTool without pipelining.}
	\label{figure:factool_timing_baseline}
    \end{subfigure}
    \vspace{1em}
    \begin{subfigure}[t]{\columnwidth}
        \includegraphics[width=\columnwidth]{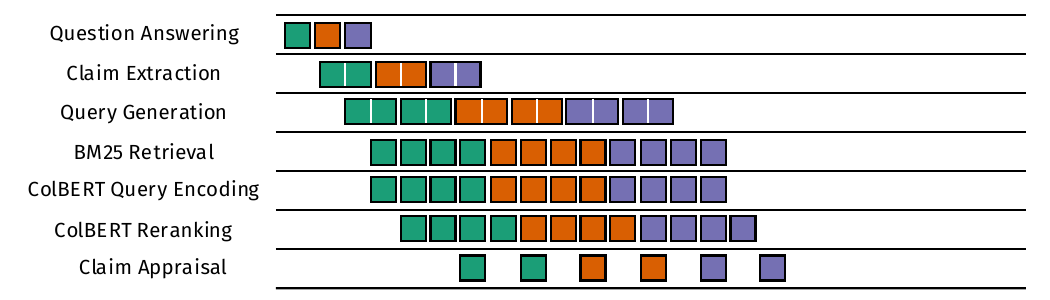}
        \caption{FacTool with pipelining.}
	\label{figure:factool_timing_streaming}
    \end{subfigure}
    \caption{Visualization of the FacTool application with and without
        streaming. Streaming reduces
end-to-end latency.}
    \label{fig:factool_timing}
\end{figure}

\subsection{Motivating Example: FacTool}
\label{sec:appsandmotivation:factool}

We illustrate some of the above properties more concretely through
a running example of the FacTool pipeline~\cite{chern2023factool}.
Figure~\ref{fig:factool_diagram} presents
the complete FacTool application.\footnote{Our instantiation of FacTool uses
a retrieval sub-graph with local components rather than the Google Search API
used in the original FacTool paper.}
This application consists of three generative language model calls
(answering the initial knowledge-based question, extracting claims to verify,
and generating search queries based on these claims),
followed by document retrieval based on the search queries,
and then ranking of those documents based on embeddings of the search queries.
A final language model call evaluates the original response and evidence together
to assess the response accuracy (claim appraisal).

FacTool is a complex pipeline. It combines data streams,
data-parallel operations,
stages that operate on different data types,
stages that can be executed in parallel with other stages,
and aggregations that require carefully tracking data sources.


The stages of FacTool accept inputs at varying granularities of text,
or input types that are not text at all.
Claim extraction takes an initial generation as input (an entire LM
output),
and search query generation takes a claim as input (a single newline).
Candidate document retrieval can process inputs token-by-token,
while the reranker accepts a set of documents and query embeddings (tensors).

Several of these stages produce multiple outputs that can be processed in parallel --
for example, claim extraction produces multiple claims per response
and search query generation can produce multiple queries per claim.
Some stages can be executed in parallel:
for example, the document retrieval and query embedding stages.

Execution can be pipelined as well: the system does not need to block until all outputs are produced to start
processing them in the downstream stage, as shown in
Figure~\ref{fig:factool_timing}.
For example, the claim extraction stage can send each claim to the search
query generation stage as soon as a claim is generated.

Finally, the computation graph imposes routing constraints as well as flexibility.
Some streams must be routed through the same stage instance (for example,
tokens corresponding to the same search query must go to the same document retrieval instance),
while elements of parallelizable streams can be routed to independent instances,
enabling fine-grained load balancing.


\subsection{Additional Compound AI Examples}
\label{sec:appsandmotivation:apps}
Many applications share these properties.
Tree of thoughts~\cite{treeofthought} is a framework for iteratively solving
challenging tasks using LMs. The core algorithm is a breadth-first-search
that uses an LM to decompose the initial problem definition into more granular
``thoughts'' which can then be repeatedly refined and evaluated through
additional LM calls to simulate structured reasoning.
Streaming helps by
enabling overlap between the generation of a thought and its subsequent
evaluation (for tasks where thoughts are generated using a single prompt).
Pipelining is especially advantageous as the time to generate a
thought and its evaluation are similar.

STORM~\cite{storm} and Wikichat~\cite{semnani2023wikichat} are two compound AI
systems that use retrieval and generation together.
STORM generates outlines and initial drafts for Wikipedia-style articles, using
a retrieval to research the given topic, and a language model to
synthesize the evidence.
Wikichat~\cite{semnani2023wikichat} is a system that refines the responses a chatbot gives
by grounding answers with data from Wikipedia.
Both systems, similar to FacTool, include opportunities to overlap
generating search queries with the search process itself.



\subsection{Design Goals}
\label{sec:appsandmotivation:abstractiongoals}
Due to the shared features of many compound AI applications, a system runtime could
automatically optimize and schedule them, given the right abstraction.
Based on the application analysis in
Section~\ref{sec:appsandmotivation:properties}, such an abstraction has 3 design
goals.

\begin{itemize}[leftmargin=*]
\item \textbf{Hierarchy of data fragmentation:} the abstraction should capture
the hierarchical structure of data passed between stages, and separate the
granularity of data a stage requires to complete (e.g., a sentence) from the
granularity it can receive (e.g., a token).

\item \textbf{Parallelism and streaming:} the abstraction should be able to
expose information about which fragments, and at what granularity, can be
parallelized across multiple instances of a stage, and which fragments, and at
what granularity, have to be streamed to the same instance.

\item \textbf{Composability:} the abstraction should support taking entire
subgraphs of stages and encapsulating them so they look like a single stage
that can be composed into larger applications. For example, a future compound
AI application should be able to take any of the applications above and
easily incorporate them as a component in a larger application.

\end{itemize}
While existing frameworks offer composability and some support for parallelism, they fail to expose
information about streaming and the inherent data hierarchy to the
runtime.
In addition to these goals,
the abstraction should provide sufficient information for a system runtime
to efficiently use resources, minimize latency of individual requests, and
maximize throughput for a fixed latency target.
User-tolerable latencies are criticals as many applications are used for
interactive tasks such as chatbots.

\section{Programming Abstractions}
\label{sec:interface}

\sys focuses on automatically parallelizing and pipelining compound AI applications.
It accomplishes this by defining a hierarchical data model and operators for
splitting, merging, parallelizing over, and transforming data items.
Pipelining and streaming are possible in many parts of compound AI systems,
and \sys provides an interface for easily expressing programs that take
the most advantage of these opportunities.

The new system pattern driving \sys is that language models often produce streams of
data, which downstream processing consumes as partial chunks.
For example, FacTool internally produces a sequence of claims,
but each individual claim can be processed as soon as it completes.
Therefore, claim processing can start before all of the claims are
generated, and claims can be processed in parallel.
Taking advantage of the pipelining and parallelization opportunities
of this pattern necessitates being able to split streams into the unit 
that subsequent processing expects, for example on textual boundaries
such as spaces and newlines.
Compound AI systems often incorporate 
standard and legacy data sources, such as embeddings, images, and documents,
whose processing performance can also improve from pipelining and
parallelization.

This section presents the two main parts of the \sys interface that enable
parallelization and pipelining: hierarchical
datatypes~\secref{sec:interface:datatypes} and a set of streaming dataflow
operators~\secref{sec:interface:dataflowoperators}. The FacTool application
is a running example to demonstrate how systems use these abstractions.

\begin{table}[t]
    \footnotesize
    \begin{tabular}{@{}l p{4.8cm}}
\toprule
{\bf Datatype} & 
\multicolumn{1}{c}{{\bf Description}} \\
\toprule

{\tt CustomObject[A]}     & 
Custom, non-text data structures. Must be serializable. \\
\midrule

{\tt Delimiter}     & 
Special text objects defined in \sys, corresponding to common delimiters used to
split natural language output (\eg newline or space). \\
\midrule

{\tt Stream[A]}     & 
Asynchronous streams generic over any of the datatypes listed here. \\
\midrule

{\tt Combined[A,B]}     & 
Tuples that join any two datatypes into a combined object. \\
\midrule

{\tt List[A]} & 
List objects generic over any of the datatypes list here. \\
\bottomrule
\end{tabular}
\caption{
    Core datatypes in the \sys interface. Stream, Combined and List objects
    are generic over any of the types here (\eg a stream can contain Combined
    objects).
}
\label{table:datatypes}
\end{table}

\subsection{Hierarchical Datatypes}
\label{sec:interface:datatypes}

An \sys \emph{stage} is a unit of data processing: it has an input type
and an output type. These types can be an individual element, such as 
a sentence or an embedding. They can also be a list, which is an ordered
sequence of a type produced or consumed as a whole, or a stream, an ordered 
sequence processed incrementally on each element in the stream.
There are also tuples, allowing two objects to be associated with one 
another (e.g., a document and the query that generated it). 
Stages operate over \sys datatypes.

Table~\ref{table:datatypes} shows these \sys datatypes.
\sys also introduces the concept of a datatype hierarchy, building on these basic datatypes. 
This hierarchy represents relationships and
dependencies between a data object and the objects generated from it.
For example, if one stage produces a list of claims and the application
splits the list into individual claims, the individual claims are
children of the source list. Similarly, if a search query produces a
list of documents, those documents are children of the query.



Capturing these hierarchical relationships makes it possible to easily
associate and manage data in a streaming and parallel system.  Unlike
bulk synchronous parallel workflows, like in Spark and other analytics
pipelines, the first output of a compound AI application may appear
before the last input is generated. Since there may be no
synchronization points, programs, and systems need to be able to
specify how a given piece of data relates to inputs in previous
stages and outputs.  The fact that generative models produce a
variable number of ``units'' of data complicates this: because one
doesn't know if the output is complete until an end-of-output
delimiter type.

A core innovation in \sys is that developers can express this datatype
hierarchy in the application. \cc{Stream} and \cc{List} objects allow
\sys to learn about this hierarchy.  Each item in a list or stream is at one level of hierarchy
below the entire stream or list.  For example, a program can specify
the output of a language model call should be split by newlines and
connect this output to a function that processes text line by line.
\sys has a library of text transformations, which automatically
perform this transformation, e.g., taking an output of paragraphs and
splitting it into lines.  Making the system aware of these data
hierarchies is not only a convenience to the programmer, but is
essential for optimizing execution.

\sys's hierarchical datatypes differ from how splittable datatypes are
used to parallelize processing in other systems: \sys's datatypes
derive from language models, which output an unknown amount of data in
a streaming fashion. The linear structure of text is such that
parallelism is represented in terms of streams: a stream of inputs can
be processed in parallel, and merged back into a stream of outputs.
Any processing elements that operate on these individual text items
must also support streams.  This contrasts with systems which split
data to parallelize processing, such as systems that process on pixels
in an image in parallel, or rows in a matrix in parallel.  




\para{Creating Stream Types in \sys}
\begin{listing}[t]
\begin{minted}[fontsize=\footnotesize,linenos,breaklines=true,xleftmargin=2.2em]{python}
class GenerateNode(AltoProcessor, Generic[X], Generic[Y]):
  @abstractmethod
  async def getUserPrompt(self, inp: X) -> str:

  @abstractmethod
  async def processOutput(self, inp: X, out: Stream[str]) -> Stream[Y]:
    # example that returns NewlineDelim
    # return alto_lib.stream_newlines(out)

  async def process(self, inp: X) -> Stream[Y]:
    prompt = await self.getUserPrompt(inp)
    textStream = await self.engine.invoke(prompt)
    return self.processOutput(textStream)
\end{minted}
    \vspace{-9pt}
\caption{\small Definition of a language model call processor in \sys which takes
    in datatype X and produces a stream of datatype Y. The function \cc{processOutput}
defines how to convert the raw text stream from the language model into the
application's needed custom object.}
    \vspace{-13pt}
\label{listing:splittingapi}
\end{listing}
 Developers use language model
stages to create stream objects
level of hierarchy below the input object.
In \sys, developers
declare language model stages with the interface shown in
Listing~\ref{listing:splittingapi}.
The language model call is
parameterized by an input datatype \cc{X} and output datatype \cc{Y}
(line 1).  The language model processor first uses a user-defined
function to transform the input object into a prompt to invoke the
model (line 9).

The raw output of the language model does not correspond to the
desired datatype \cc{Y}, so the processor must convert the raw stream
of text into a stream of \cc{Y}.  As this is application-specific,
developers must implement the \cc{processOutput} function (line 6) to
define this mapping.  Developers can use library functions to
automatically produce streams of \cc{Delimiter} items that correspond to
common ways to split text, such as splitting by newline or space.

While the current applications we consider only have instances of
streams created from language model calls, the interface supports any
processing stage that takes one input and creates many outputs.  \sys
also provides ways to convert \cc{List} objects into streams.  A RAG
application, for example, could include a retriever that produces a
set of top-k documents, but does post-processing on each document to
summarize each document in context of the query.
In \sys, developers can also split \cc{Delimiter}
objects into streams of smaller \cc{Delimiter} items (\eg splitting a line into
words).

\begin{table}[]
    \footnotesize
    \begin{tabular}{lll}
\toprule
\bf Operator & 
\multicolumn{1}{c}{\bf Inputs} & 
\multicolumn{1}{c}{\bf Outputs} \\ 
\toprule

{\tt call\_stage} & 
\begin{tabular}[c]{@{}l@{}}
    {\tt f: Stage[A->B]}\\
    {\tt inp: A}
\end{tabular} & 
{\tt B}\\ 
\midrule

{\tt call\_lm} & 
\begin{tabular}[c]{@{}l@{}}
    {\tt g: GenerateNode[A,B]}\\
    {\tt inp: A}
\end{tabular} & 
{\tt Stream[B]}\\ 
\midrule

{\tt pcombine} & 
\begin{tabular}[c]{@{}l@{}}
    {\tt left: A}\\
    {\tt right: B}
\end{tabular} & 
{\tt Combined[A, B]}\\ 
\midrule

{\tt split\_text} & 
{\tt inp: Delimiter} &
{\tt Stream[Delimiter]}\\ 
\midrule

{\tt pmap}     & 
\begin{tabular}[c]{@{}l@{}}
    {\tt func: f[A -> B]}\\
    {\tt inputs: Stream[A]}
\end{tabular} &
{\tt Stream[B]}\\
\bottomrule
\end{tabular}

\caption{Operators in the \sys dataflow interface.
    \cc{call\_stage} and \cc{call\_lm} invoke processing stages.
    \cc{pcombine} and \cc{split\_text} enforce type consistency between stages by
    joining separate objects or splitting a text item by a delimiter.
    \cc{pmap} expresses parallelization in \sys: \sys will call
the specified function in parallel on the stream and aggregate the outputs.
\sys also includes support for conditional control flow, which we omit for
brevity.}
    \vspace{-10pt}
\label{table:operators}
\end{table}

\subsection{Dataflow Operator Interface}
\label{sec:interface:dataflowoperators}

The \sys dataflow interface allows developers to connect stages into
complex compound AI pipelines while specifying streaming and
parallelization semantics. The interface is designed to allow easy
composition. Developers write individual stages, connect them into
larger units of functionality using the operators, and then invoke
those larger units as functions. The entire FacTool pipeline, for
example, can be invoked as a function that takes knowledge-based
question as input, and outputs a list of combined \cc{(Newline,}
\cc{List[VerifiedClaim])}
tuples, consisting of factual claims and information about whether they are
supported by evidence.

\subsubsection{Operators. }
Table~\ref{table:operators} defines the various ways developers can
process datatypes and connect stages together.  With these operators,
it is possible to connect different stages where the first stage's
output does not match the next stage's input.  There are three categories of operators: (i) processing operators which
call individual stages on datatypes, (ii) data manipulation operators which
manipulate datatypes before passing them into stages, and
(iii) parallelization/aggregation operators. We describe each in turn.

\para{Processing Operators}
To process data, developers can use the \cc{call\_stage} and 
\cc{call\_lm} operators.  These operators take as input whatever
the stage takes as input (which the developer must declare), and
produce what the stage outputs.  Because applications commonly create
streams from language model outputs, the \cc{call\_lm} operator
outputs a \cc{Stream} datatype.

\para{Data Manipulation}
\sys includes two operators that manipulate data. First, the \cc{split\_text} operator modifies how the data is sent and consumed to
allow text to be consumed at a finer granularity.  For example, a
developer could call \cc{split\_text} on a newline in a language
model output, to further split it into words.

Second, the \cc{pcombine} operator allows developers to create combined
(tuple) objects.  This is useful for cases where application stages
receive data from different parent stages; the parent stages can
work in parallel.

\para{Parallelization/Aggregation}
The \cc{pmap} operator defines parallelization and aggregation in \sys
applications.  The \cc{pmap} operator takes a stream and calls a
function over each stream item, returning a stream of the function
outputs.  As each stream item finishes, the \cc{pmap} operator
reconstitutes the stream in the original order and sends the output to
the next part of the program in a streaming fashion.  The function
\cc{pmap} is called over a function which can further contain arbitrary \sys operators,
potentially even other \cc{pmap}s.  \cc{pmap} allows \sys applications
to compose: a future application could call an existing
application in parallel over a data stream. \cc{pmap} is similar in
spirit to a map task in MapReduce~\cite{mr}, but the ``map''
task can itself contain a multi-stage application (with inner dataflow
operators).  Additionally, data {\em streams} into and out of the
parallel function calls.

\begin{listing}[t!]
\begin{minted}[fontsize=\footnotesize,linenos,breaklines=true,xleftmargin=2.2em]{python}
def processClaim(claim: Text[NewlineDelim]) ->
CustomObject[VerifiedClaim]:
  sqgStream = call_lm(sqg_generate, claim)
  docsStream = pmap(sqgStream, search)
  claim_with_evidence = pcombine(claim, docStream)
  return call_stage(verifier, claims_with_evidence)

def search(query: Text[NewlineDelim]) -> CustomObject[Docs]:
    words_stream = query.split_text(WordDelim)
    embeddings = call_stage(query_encoder, query)
    candidates = call_stage(bm25, words_stream)
    combined = pcombine(embeddings, candidates)
    return call_stage(reranker, combined)
\end{minted}
    \vspace{-5pt}
\caption{Example of all the operators in the FacTool
application, to parallelize processing over a stream of search queries. Though
not shown, note all functions are asynchronous.}
    \label{listing:pmap}
    \vspace{-10pt}
\end{listing}

\paragraph{Example from FacTool. }
Listing~\ref{listing:pmap} shows the part of the FacTool application implementation, which uses all \sys operators.
It focuses on processing each claim.
The application generates a stream of search queries, calls a search function
in parallel over the stream, and passes the output to a final verifier
stage.
The inner search function, which operates at the granularity of a query, represents a subgraph of stages.
The search function passes the query into BM25, after splitting it into
words, to gather candidate documents.
Concurrently, it generates embeddings for the search query.
It passes both the embeddings and candidate documents to the reranker stage, as
a combined object, to gather the final top-k documents.

\subsection{Limitations}
Writing applications in \sys requires more boilerplate code than is presented in the previous section.
The boilerplate code helps \sys run in a distributed environment where
stages are located in different physical processes, and a central forwarder
forwards data between stages (following the graph structure).
Application developers thus must:
\begin{enumerate}
    \item Declare queues between stages, and extra send and receive operators to pass data along the graph.
    \item Declare which queues represent \cc{pmap} entry and exit points. This allows the engine to handle parallel calls.
    \item Invoke an extra \cc{aggregate} operator that collects data from a network queue that is part of the same stream (\eg after a pmap), and collect it into a stream object.
\end{enumerate}

\section{Runtime}
\label{sec:runtime}

\sys has a distributed runtime for streaming compound AI
dataflows. The runtime automates scheduling pipelining
and parallelism while ensuring that every data element 
arrives at the correct stage instance in the correct order,
enforcing and managing the dynamic data hierarchy a program creates.

As a concrete example, consider the search query
generation stage of the FacTool pipeline. This
stage generates a stream of search queries. Each query is split into
tokens, which are streamed to document retrieval. While
each query's tokens must be sent to a single instance,
different queries can be sent to different instances. At
the same time, each query is sent to query encoding;
in this case, queries have to be sent in entirety (streaming
their tokens has no benefit), but separate queries can
be sent to different encoding instances.

The results of these stages -- documents and embedding
tensors -- are then combined and processed together. This
requires that the runtime can associate a particular document
and its associated embedding with the 
particular query. It must do this while streaming partial results
and correctly scheduling them across multiple instances of stages.

\subsection{Execution is a Routing and Ordering Problem}

\sys programs impose two requirements on how the runtime routes data and
places computations. First, if a stream is parallelized, later stages that operate on the output stream sequentially
expect it reconstituted into the original order. In the FacTool example above,
this is necessary to aggregate all the documents across all the generated search
queries (line 4).

Second, stages can be either stateful or stateless. Stateful stages maintain
incremental results on elements of a stream and must see every element of the stream.
The document retrieval stage, for example, can accept streamed tokens, but requires that
every token associated with a search query is sent to the same instance; while it can
incrementally compute on tokens as they arrive, it needs all of them to produce 
documents. Stateless stages, in contrast, process each element of a stream
individually. Stateless stages can parallelize a stream across instances, unlike stateful stages.



Meeting these two requirements requires tracking both the index of an element within a stream
as well as an identifier of the source that generated the stream. In early
versions of \sys, programmers managed this bookkeeping manually (within
program data structures), but it was
painful and error prone. Furthermore, it made composition very difficult as
the metadata structure was embedded within the program's data structures.
Therefore, \sys automatically tracks the data hierarchy from program structure using \emph{nested ancestry}.

\subsection{Nested Ancestry}

\begin{figure}[t]
  \centering
  \includegraphics[width=\columnwidth]{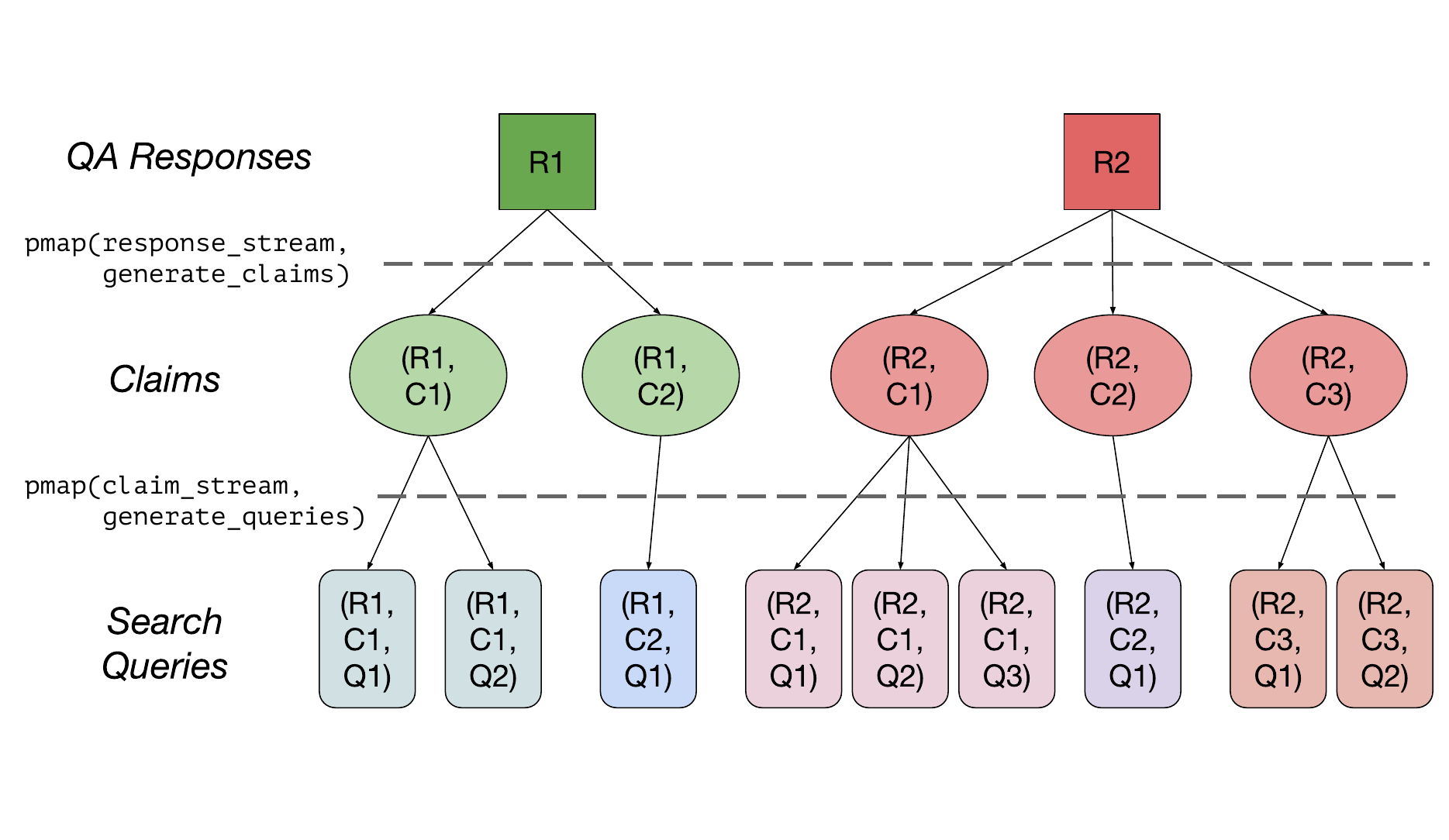}
  \caption{Example of nested ancestry tags for the first three stages in the FacTool pipeline.
    As \code{pmap} calls are made, levels of ancestry are added, counting both the level of parallel nesting
    as well as the index of elements output by each input element.}
  \label{fig:ancestry}
\end{figure}

Nested ancestry is metadata that accompanies every data element in \sys.
It is a stack of tags that tracks the index of the element in the stream
as well as the parent elements that generated this stream element.
Figure~\ref{fig:ancestry} illustrates nested ancestry tags for stages 
in FacTool.  A search query is generated from a claim,
which was in turn generated from the response to a top-level request (knowledge-based question).
Hence, each search query is tagged with the index of the query, the ID of the source claim, and the ID of the request that generated the claim.
These tags allow the reranking stage to aggregate results from the document retrieval and embedding 
tensor generation stages corresponding to the same claim and the same query.

\sys tracks ancestry automatically throughout query execution to take the burden off 
the developer to do so. Tracking ancestry on every stage call is too expensive --
this would amount to keeping track of the entire computation graph and
provenance of every output.

The key insight that makes the \sys scalable is that the only operation that requires inserting
an ancestry tag is \code{pmap}. This operator indicates that the input stream elements are independent 
and can be operated on in parallel.
The primary purpose of ancestry tags is to allow for aggregation after a parallel call.
Hence, it is sufficient to add a level of ancestry on \code{pmap} entry
and remove a level of ancestry on \code{pmap} exit. In this way, nested ancestry is
tightly coupled with program structure: it represents a call stack of {\tt pmap} operators,
which can be arbitrarily nested through program composition.

Levels of the data hierarchy besides \code{pmap} do not require explicit tracking because
they are implicitly represented by ordering and program structure. Because \sys
uses in-order byte streams (e.g., TCP) as its underlying transport, stream elements sent
to a single instance of a stage arrive in order. The program structure itself sets
up the levels of data hierarchy: any given connection between two stages carries a single
type from a single position in the program.

\paragraph{Composability.} Compound AI systems are built from components reused across many different
query pipelines (for example, consider document retrieval, which is a core component for many AI applications
beyond FacTool). FacTool, for example, could be a component of a larger system.
Because \sys tracks ancestry automatically,
a developer can freely reuse the same implementation of a subgraph in any computation graph, as \sys will correctly handle metadata tracking.

\subsection{Routing with Ancestry}
A \code{pmap} entry indicates pushing an ancestry tag and a parallelized execution.
A \code{pmap} call exit pops an ancestry tag and aggregates over the outputs generated in the call.
This aggregation requires routing the relevant elements to the same physical instance of
the next stage.

Nested ancestry enables \sys to easily route all stream elements to the right instance of 
a stage. If there are multiple instances of a stage, all elements with the same set of
ancestry tags will be routed to the same instance. In this way, routing in \sys is similar to
Equal Cost Multipath Routing (ECMP)~\cite{ecmp}, except that instead of routing based 
on packet headers, it routes based on ancestry tags. Also, because stream elements are generated
at a much lower rate than IP packets, the decision of which stage to forward a new ``flow'' of
tags to can be dynamic and based on load rather than hashing.



\section{Implementation}
\label{alto:impl}
Our prototype implementation of \sys includes a Python application
library to construct application graphs and implement applications
(\textasciitilde 4500 LOC), and a distributed controller written in Rust (\textasciitilde 5500 LOC).
Developing an application in \sys requires writing separate microservices
which contain stage logic and calling the \sys operators via the Python
library.
Running the application requires specifying the application graph to the
controller, by specifying where queues are,
which queues
represent \cc{pmap} entry/exit points, and which queues represent data that needs to
be aggregated.
The controller spawns each service and forwards data along queues between
stages, choosing replicas using the routing rules and queue length
monitoring.
The core operators are implemented in the Python interface,
but the engine handles \cc{pmap} tracking.

\section{Evaluation}
\label{alto:eval}
Our evaluation finds that \sys outperforms or matches hand-optimized implementations of various complex compound AI systems, specifically:
\begin{enumerate}
    \item \sys enables intra-query parallelism and streaming,
        allowing \sys to have lower latency and higher throughput than baselines that cannot distribute and stream work within a query.
    \item \sys does not introduce application overhead; in cases where the application itself does not benefit from
        pipelined execution, \sys matches the performance of hand-optimized
        baselines.
    \item \sys achieves benefits both due to its ability to send work early
        and distribute independent work across different workers.
\end{enumerate}

\subsection{Methodology}
\label{alto:eval:methods}
This section describes our hardware, evaluated applications, baselines,
and metrics.

\subsubsection{Setup}\hfill
\label{alto:eval:methods:setup}

\para{Hardware}
We evaluate FacTool~\cite{chern2023factool},
Tree-of-Thought~\cite{treeofthought}, and
WikiChat~\cite{semnani2023wikichat} on a single NVIDIA HGX node with eight 80 GB A100-SXM GPUs, a 256 core AMD EPYC 7763 CPU and 1056 GB of RAM.
We evaluate the fourth application (STORM~\cite{storm}) on a machine inside a SLURM
cluster, with 1 Quadro RTX 6000 GPUs (24 GB) and 7 CPUs (Intel Xeon Platinum
8268 CPU, 2.90GHz), and 375 GB RAM.

\para{Metrics}
For all applications, a client process sends requests at a fixed rate and
we measure the latency distribution of the responses.
Each request is a \emph{program request}, so latency refers to end-to-end latency \emph{across the entire compound AI
application}, rather than the latency of individual component calls.
For FacTool and WikiChat, we show throughput vs latency curves, and report median and 99th
percentile tail latency.
For Tree-of-Thought and STORM we study latency distributions
where all systems meet the offered load.

\subsubsection{Applications, Workloads}
\hfill
\label{alto:eval:methods:apps}

\para{Common Subcomponents} We instantiate four compound AI applications with a local model serving
engine (using open-source weights) and local search components, unless otherwise
specified.
Original implementations of these applications use API services (for both
retrieval and generation), but \sys
currently targets local deployments, so we evaluated the applications with
different components.
\sys can be made to work with API services, but using local components
allows for a more controlled performance comparison.

Across all applications, we use vLLM as the language model engine, version
0.63.post1.
For Tree-of-Thought, prefix caching is turned on.
For FacTool and STORM, which contain search retrieval sub-applications, we
instantiate a
3-stage retrieve-and-rerank search
pipeline~\cite{nogueira2019multi}, using a BM25~\cite{bm25} retriever to gather
initial candidate documents, a
ColBERT~\cite{santhanam2022colbertv2, santhanam2022plaid} query encoder to turn the search query
into embeddings, and a ColBERT reranker to rerank the candidates with the
embeddings.
For WikiChat, we use a single-stage ColBERT retriever for search.
All retriever pipelines use an index built over Wikipedia, unless
otherwise specified.

\para{FacTool}
FacTool represents a workload that benefits from \sys's automatic
pipelining of execution, as it involves processing at three nested levels of
hierarchy: for each request, each claim, and each search query.
We instantiate the FacTool application (as described in prior sections) with
Open-Orca's Mistral-7B model for all language model calls and the 3-stage
retrieve-and-rerank retriever subsystem.
Our evaluations use SQuAD~\cite{rajpurkar2016squad} queries as the input
data, and each data point involves feeding in
queries for 12 minutes at the specified offered load.
FacTool involves 4 language model prompts -- one for creating the initial
response, one for extracting claims, one for extracting search queries, and
one for verifying the claims with the retrieved evidence.
The \sys implementation combines the three shorter prompts into one LM service
and keeps the claim appraisal prompt separate.

\para{Tree-Of-Thought}
Tree-of-Thought benefits from \sys's automatic parallel processing over lists
and streaming of language model output.
Tree-of-Thought is a reasoning framework for language models, with three logical
stages: thought proposal, thought evaluation, and thought selection.
The framework uses an LM to propose ``thoughts'' that help incrementally solve
a problem.
An evaluator LM scores all the proposed thoughts, and the selection phase
chooses the top-K thoughts.
This 3-stage process proceeds in a loop for 3 iterations, resulting in three
levels of parallel nesting: per request, per parent thought, per proposal.
The \sys implementation requires an additional aggregation stage, due to an
engineering limitation in the interface where it cannot aggregate streams of
streams correctly without intermediate computation at each aggregation.
We evaluate the Game24 problem, and use 32 of the queries used in the
original paper.
Each game results in 100s of model calls; each parent thought proposes 30-50
children thoughts, which are each scored three times each, for three iterations.
Each data point in the evaluation represents the latency distribution of sending
30 queries at the specified offered load.

\para{WikiChat}
WikiChat uses data from Wikipedia to improve the quality of language model
responses in chatbot scenarios.
It involves two concurrent processing paths: generating an initial response and
extracting claims from the response (similar to FacTool), and finding data from
Wikipedia relevant to the query.
Once it has the extracted claims and relevant data, it generates a draft
response and optionally refines it before returning it to the user.
We evaluate the complete combine pipeline, using queries from
the SQuAD dataset. 

The \sys implementation allows parallel processing over the stream of claims,
which are produced from partial outputs from the language model. In parallel,
The LM also generates search queries based on the user input. Retrieval is then
performed on the claims and the search queries; the results are then verified, summarized
using LM stages. Finally a response is drafted and then refined.
Each datapoint presented is the result of sending queries, at a specified rate for 5 minutes.
Our instantiation of WikiChat uses the Qwen2.5-32B-Instruct model and the ColBERT Wikipedia index.

\para{STORM}
STORM is a pipeline that helps editors with pre-writing tasks, specifically in
the context of generating outlines and drafts of Wikipedia-style articles from
scratch.
It represents the most complex application integration in \sys.
The pipeline generates an initial outline, identifies relevant perspectives to
write the article using the table of contents from existing related Wikipedia
articles, and searches for relevant evidence using these perspectives.
It then refines the outline using the evidence.

We implement STORM until the refine outline stage and omit the step of
generating the final draft.
Our instantiation of STORM uses the Mistral7B model, a web API to scrape 
the table of contents from existing Wikipedia articles, and the 3-stage retrieve-and-rerank pipeline used
for FacTool for search.
Due to limitations in the SLURM environment used to evaluate STORM, the search
pipeline uses a smaller test index based on the LoTTe~\cite{santhanam2022colbertv2} dataset.

STORM contains various opportunities for parallelizing on streams, but we
expect no speedups, because either these opportunities represent a
small percentage of overall processing time, or the time to process stream
inputs largely outweighs the time to generate all stream items.

\begin{table}[t]
    \footnotesize
    \begin{tabular}{l r}
\toprule
Application & Stages (and replicas) in \sys Implementation \\ 
\toprule

\textbf{FacTool} &
\begin{tabular}[r]{@{}r l@{}}
    Driver & 4 (CPU)\\
    Initial LM Stages & 3 (GPU, CPU)\\
    BM25 & 4 (CPU)\\
    ColBERT Query Encoder & 4 (GPU, CPU)\\
    Claim Appraisal LM & 4 (GPU, CPU)
\end{tabular}\\
\midrule

\textbf{Tree-of-Thought} &
\begin{tabular}[r]{@{}r l@{}}
    Driver & 8 (CPU)\\
    LM Stages & 8 (GPU, CPU)\\
    Aggregator & 8 (CPU)\\
    Thought Selector & 8 (CPU)
\end{tabular}\\
\midrule

\textbf{STORM} &
\begin{tabular}[r]{@{}r l@{}}
    Driver & 1 (CPU)\\
    LM Stage and Query Encoder & 1 (GPU, CPU)\\
    BM25 & 1 (CPU)\\
    Reranker & 1 (CPU)
\end{tabular}\\
\midrule

\textbf{WikiChat} &
        \begin{tabular}[r]{@{}r l@{\hspace{0.7cm}}}
    Driver & 4 (CPU)\\
    LM Stages & 4 (GPU)\\
    ColBERT Retriever & 4 (CPU)\\
\end{tabular}\\
  
\bottomrule
\end{tabular}

\caption{Summary of applications evaluated, in terms of components and
resource allocations in the \sys implementation.}
\label{table:apps}
\end{table}

\subsubsection{Evaluated Baselines}
\label{alto:eval:methods:baselines}
We compare \sys implementations of applications to versions written in
LangGraph~\cite{chase2022langgraph}, a framework for AI agents.
These versions are hand-optimized, parallelizing on partial outputs
manually with Python's async API.
For example, the FacTool LangGraph implementation contains 3 separate ``graphs''
for request processing, claim processing, and search query processing, and
manually pipelines data to each sub-graph.
For FacTool and Tree-of-Thought, we additionally compare to a version using
\sys's orchestration, without any pipelining (``Microservices without
Pipelining'').

\para{Distribution}The experiments compare a version of \sys, where each stage is replicated
independently over the available resources, to replicating the \emph{entire
LangGraph process}.
Table~\ref{table:apps} shows how each service is replicated.
The \sys version has access to multiple GPUs in any single query, while the
LangGraph version does not.
Nested ancestry allows the execution engine to distribute work over
separate processes while aggregating data correctly.
The microservices baseline shares the same replication factors.
In all applications except STORM (which is run on a SLURM cluster), we use
\cc{taskset} to pin application stages to cores, and ensure each baseline is
given an equal number of cores overall.

\subsection{End-to-End Results}
\label{alto:eval:e2e}
\begin{figure}[t!]
     \centering
     \begin{subfigure}[t]{0.90\columnwidth}
         \centering
         \includegraphics[width=\linewidth]{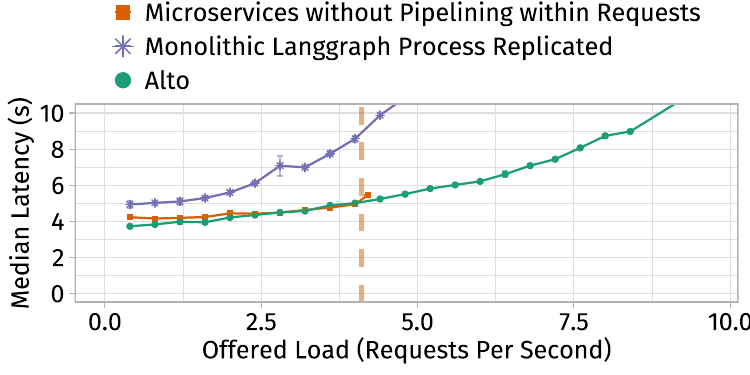}
         \caption{Throughput-median latency tradeoff for \sys compared to
             baselines. \sys can be used to achieve the same
         median latency of 6 seconds at 2.3\X higher offered load, compared
         to Langgraph.The
     microservices baseline fails to meet the achieved load after rate 4.2.}
         \label{fig:factool-e2e-median}
         \hspace{0.05cm}
     \end{subfigure}
     \vfill 
     
     \begin{subfigure}[t]{0.90\columnwidth}
         \centering
         \includegraphics[width=\linewidth]{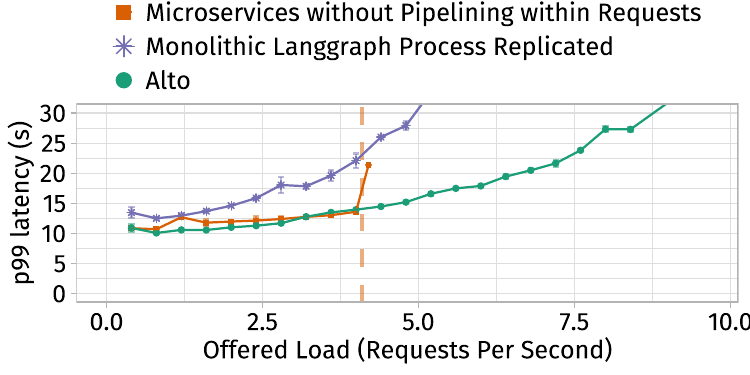}
         \label{fig:factool-e2e-p99}
         \caption{Throughput-p99 latency tradeoff for \sys compared
             to baselines. \sys can be used to achieve the same
         tail latency of 16 seconds at 2\X higher offered load.}
     \end{subfigure}
     \vfill
     \caption{FacTool End-to-End Throughput-Latency Tradeoff}
     \label{fig:factool-e2e}
     \hspace{0.11cm}
\end{figure}

\begin{figure}[t!]
     \centering
         \centering
         \includegraphics[width=\columnwidth]{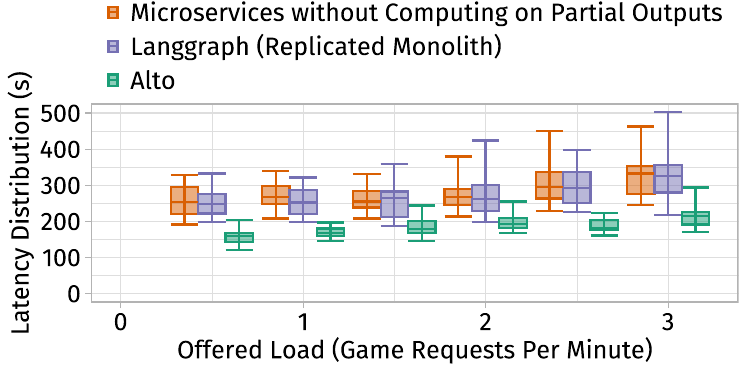}
         \caption{Throughput-latency for \sys compared to baselines, on the
             Tree-of-Thought application, showing latency distributions at
             various offered load.
             Whiskers show p5 and p99 (effectively max, as there
             are  30 requests)
             latencies, while the box shows p25, p50, and 75
         latencies. LangGraph has 1.35-1.61\X higher median latency than
         \sys, while the no pipelining baseline has 1.39-1.69\X higher
         latency than \sys. \sys lowers latency because it can start work early
     and distribute work across workers.}
         \label{fig:tot}
\end{figure}

\begin{figure}[t!]
    \centering
        \includegraphics[width=\linewidth]{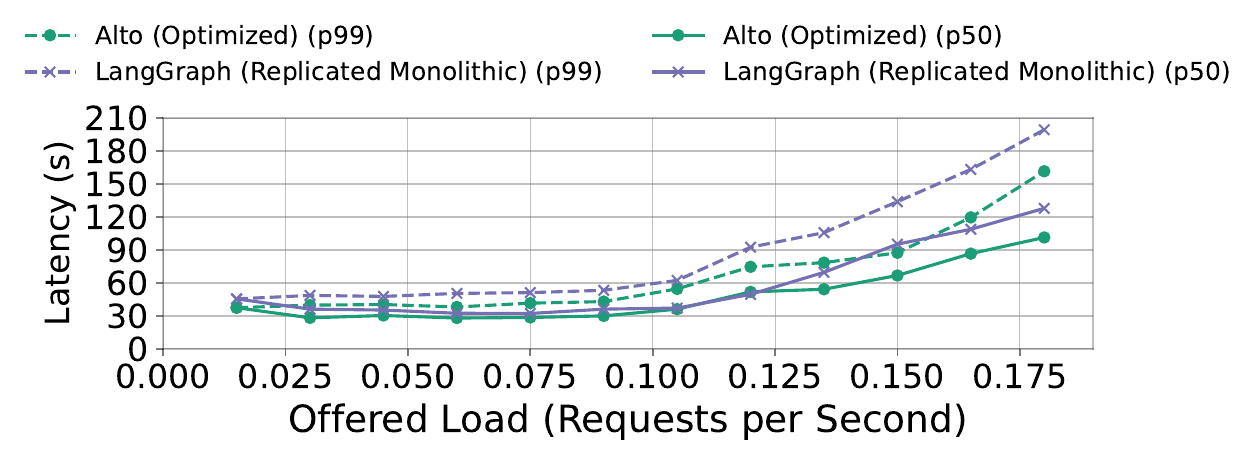}
        \caption{Throughput-latency tradeoff for \sys compared to baseline.
        \sys can be used to achieve the lower median and tail latencies
        for the same offered load, compared to Langgraph. }
    \label{fig:wikichat-e2e}
\end{figure}

For each workload and application described in \secref{alto:eval:methods:apps},
Figures~\ref{fig:factool-e2e},~\ref{fig:tot}, ~\ref{fig:wikichat-e2e} and Table~\ref{table:storm} show the performance of \sys compared to
a replicated monolithic LangGraph baseline, and for FacTool and Tree-of-Thought,
compared to the microservices baseline without pipelining.

\para{Summary}
For FacTool, LangGraph has 1.26\X and 1.25\X higher median and tail
latency at the lowest offered load.
\sys also achieves 2.3\X higher throughput than LangGraph at the same median latency
SLO of 6 seconds.
For Tree-of-Thought, at low load, LangGraph has 1.35\X to 1.69\X higher
median latency than \sys.
For STORM, at low load, \sys and LangGraph achieve about the same median
latency. For WikiChat, LangGraph achieves a 1.13\X higher median latency and 
1.18\X higher tail latency for the same offered loads.

FacTool and Tree-of-Thought benefit from the ability for pipeline execution
and parallelize processing over distributed workers.
Wikichat benefits as well, but to a lesser extent.
As expected, STORM does not particularly benefit from pipelining execution
and \sys performs similarly to the baseline.

\para{FacTool}
Figure~\ref{fig:factool-e2e} shows the throughput latency tradeoff for \sys
against the baselines, for data points where offered load is within 95\% of
the achieved load.
\sys outperforms the monolithic LangGraph baseline in terms of latency at low
load: LangGraph has 1.26\X higher median and 1.25\X higher tail latency at the
lowest offered load of 0.4 requests per second.  
In terms of throughput, \sys achieves 2.3\X higher throughput for the same
median latency SLO of 6 seconds, and 2\X higher throughput at the tail latency
SLO of 16s.
\sys achieves about the same latency as the microservices baseline until
the offered load of 4.1 requests per second (the dashed orange line), after which the baseline does
not meet the offered load.

At low load, \sys and the microservices baseline achieve lower latency than the
replicated LangGraph baseline.
Serving the application as microservices, where an individual query can flow
through different GPUs benefits execution, both because
there is less load on the 1 GPU used for the request, and because
microservices allow for uneven and independent resource allocation across
processing stages.
The resource allocation used for \sys places claim appraisal on four separate
GPUs, and uses 3 GPUs for the initial language model stages.

At high load, \sys outperforms both LangGraph and the microservices baseline.
The LangGraph baseline is structured to pipeline execution manually, but
within a single process attached to a single GPU.
The microservices baseline does all the work for a particular stage (\eg
creates search queries for all claims) before submitting work to the next
stage, which creates bursts of load.
In contrast, \sys can both pipeline execution and distribute independent work
across processes.
As Section~\ref{alto:eval:breakdown} explains, we found that most of the
benefit in \sys comes from pipelining execution of partial outputs; whether
or not language model outputs are sent as they are produced has minimal effect
on the aggregate latency statistics.

\emph{\textbf{Takeaway: }On the FacTool workload, \sys outperforms both
    baselines because it can pipeline execution across distributed processes, which is made possible by nested ancestry.}

\para{Tree-Of-Thought}
Figure~\ref{fig:tot} shows the latency statistics for \sys, compared to
LangGraph and the microservices baseline, at low loads where all systems meet
the offered load.
The distributions show data for the end-to-end latency of sending 30
requests (the middle of 32 requests sent
by the client); because each request takes on the order of minutes, we sampled
fewer requests.
The baselines achieve 1.35\X
to 1.69\X higher median latency than \sys at all loads.

In this application, the benefit comes purely from pipelining execution and
distributing work across independent workers, balancing the work across
time and space.
The first stage (thought proposal) generates a
variable number of thoughts the next stage evaluates by
using an LM to score each thought.
In \sys, thought evaluation starts as each thought is generated and all
evaluations run independently across different workers.
Both baselines wait until all thoughts are generated to start evaluation.
Additionally, both baselines perform all evaluations within a single
query on the same worker, leading to bursts of work every time thoughts are
generated.

\emph{\textbf{Takeaway}: On the Tree-of-Thought workload, \sys outperforms both
baselines because it can pipeline execution, and balance load in the query
across space and time.}

\begin{table}[t]
    \footnotesize
    \begin{tabular}{l r r r r}
\toprule
\textbf{System} (rate) & p25 (s) &  p50 (s) & p75 (s) & p95 (s)\\
\toprule

\textbf{LangGraph} (rate=0.5 rpm) & 83 & 92 & 103 & 123 \\
\midrule
\textbf{\sys} (rate=0.5 rpm)  & 81 & 87 & 103 & 131 \\
\midrule
\textbf{LangGraph} (rate=1.0 rpm) & 78 & 88 & 110 & 192 \\
\midrule
\textbf{\sys} (rate=1.0 rpm)  & 68 & 76 & 89 & 113\\
\bottomrule
\end{tabular}

\caption{Latency distribution of \sys and LangGraph running the STORM
application, at an offered load of 0.5 and 1 requests per minute, for 39
requests (excluding the first out of 40), each using 1 GPU. \sys has a
similar latency distribution as this application does not benefit from pipelined
execution, and there is no opportunity to distribute with 1 GPU.}
\label{table:storm}
\end{table}

\para{WikiChat}
Figure \ref{fig:wikichat-e2e} shows the latency statistics for \sys on the WikiChat application,
comparing the \sys implementation to a monolithic replicated LangGraph baseline.
LangGraph achieves at 12.7\% or higher median latency and an 18.3\% or higher tail
latency at all loads.

The benefits of \sys are significant but more limited compared to previous applications
because only a signle stage (claim generation) produces partial outputs from the language model
which can be parallelized. Due to the streaming of partial outputs, \sys
performs retrieval based on the claims
earlier than the baseline. The latency results for \sys benefit
due to load balancing across replicas for each stage.

\para{Storm}
Table~\ref{table:storm} shows latency statistics for \sys on the STORM
application compared to LangGraph, when sending 40 requests for 1.0 or 0.5
requests/min, recording the latency of everything but the first request.
\sys and LangGraph achieve about the same overall latency at both rates.
This is expected: there are only two opportunities to pipeline execution (both
with 1 level of nesting).
The first opportunity represents a small portion of the overall processing time:
retrieving the table of contents (using a web scraper) from related
Wikipedia articles, as the names of related articles are generated.
The second opportunity represents a case where the 2nd pipeline stage takes much longer than the 1st stage.
An agent performs research with a ``perspective'' on a stream of generated
perspectives.
This processing (per perspective) involves three language model generations
(one with document processing), and the entire retrieve-and-rerank
pipelines, in a loop up to 3 times.
Due to restrictions, we performed this evaluation in a different cluster and
could not evaluate whether \sys provides benefits in a distributed GPU
scenario.

\subsection{\sys's techniques}
\label{alto:eval:breakdown}
\sys allows applications to automatically pipeline execution in a
distributed environment.
This section shows how \sys improves performance compared to the
no-pipelining scenario, and breaks down whether specific features in \sys
(directly sending output early from the language model, and fine-grained
parallelization) help with performance.

\begin{figure}[t]
    \centering
    \begin{subfigure}[t]{\columnwidth}
        \includegraphics[width=\columnwidth]{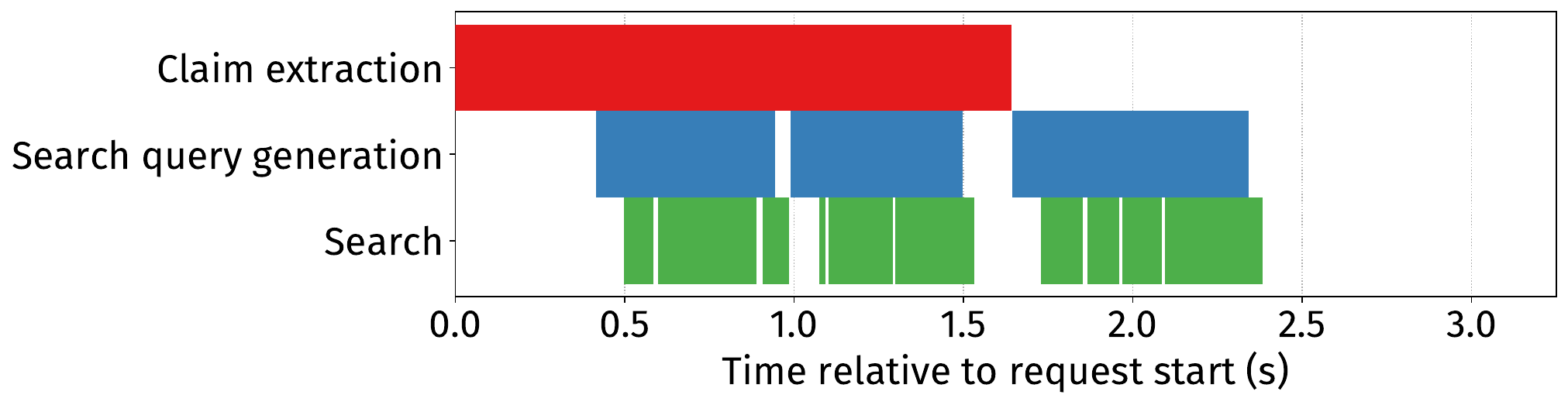}
        \caption{FacTool execution with pipelining.}
	    \label{fig:factool_visa}
    \end{subfigure}
    \vspace{1em}
    \begin{subfigure}[t]{\columnwidth}
        \includegraphics[width=\columnwidth]{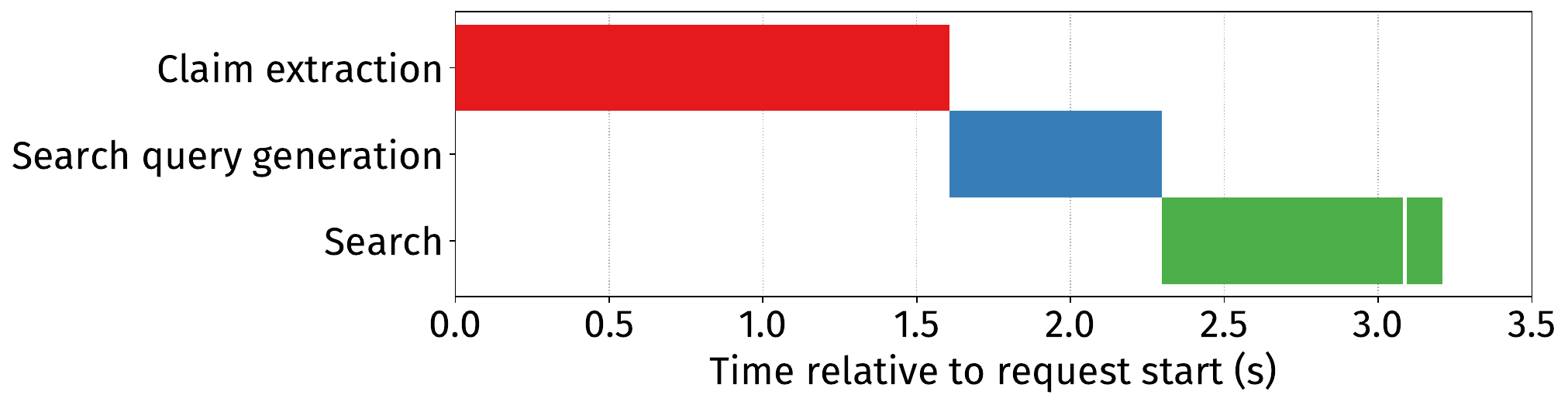}
        \caption{FacTool execution without pipelining.}
	\label{fig:factool_visb}
    \end{subfigure}
    \vspace{-10pt}
    \caption{Visualization of the FacTool application with and without
        pipelining, focusing on extracting claims, extracting search
    queries, and the search sub-graph.}
    \label{fig:factool_vis}
\end{figure}

\begin{figure}[ht!]
    \centering
    \begin{subfigure}[t]{\columnwidth}
        \includegraphics[width=\columnwidth]{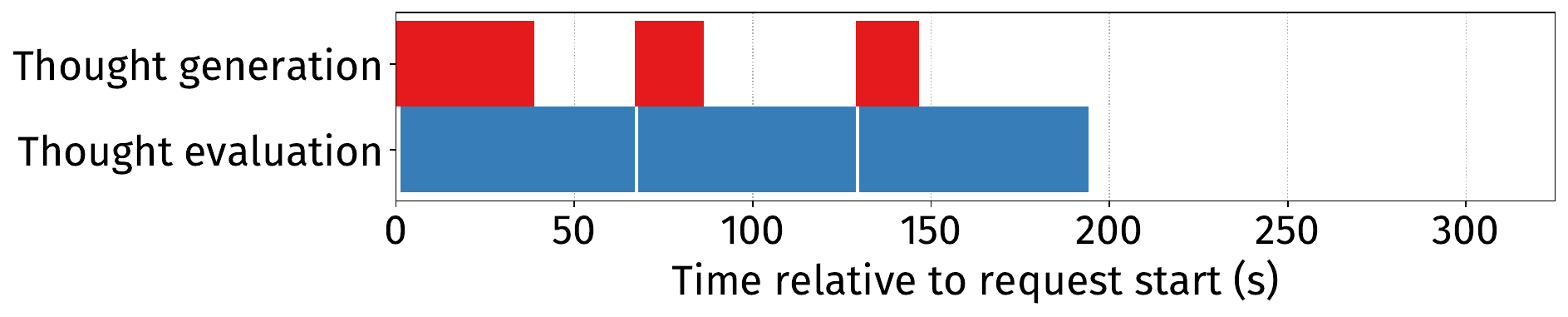}
        \caption{Tree-of-thought with pipelining.}
	\label{fig:tot_visa}
    \end{subfigure}
    \vspace{1em}
    \begin{subfigure}[t]{\columnwidth}
        \includegraphics[width=\columnwidth]{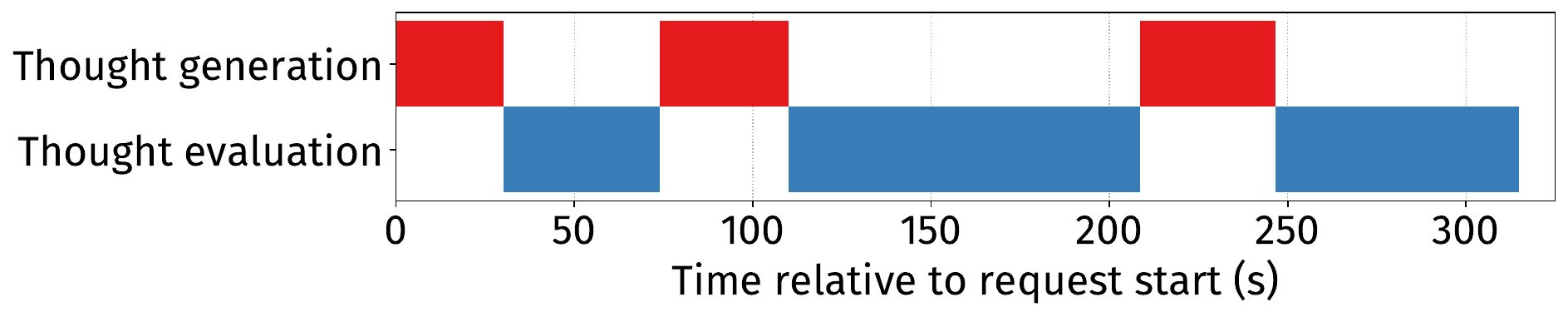}
        \caption{Tree-of-thought without pipelining.}
	\label{fig:tot_visb}
    \end{subfigure}
    \vspace{-10pt}
    \caption{Visualization of the Tree-of-Thought application with and without pipelining, across the thought proposal and thought evaluation
    stages.}
    \label{fig:tot_vis}
\end{figure}

\subsubsection{Visually Showing Pipelining}
Figure~\ref{fig:factool_vis} shows how pipelining reduces latency in
FacTool application, zooming in on the claim extraction, search query
generation, search, and claim appraisal phases, within a single query, for a
particular request, at a low offered load (0.4 rps).
Figure~\ref{fig:tot_vis} does the same for Tree-of-Thought, across its two
language model stages, at a low offered load (0.25 requests per minute).
Figures~\ref{fig:factool_visa} and ~\ref{fig:tot_visa} refer to \sys, while
Figure~\ref{fig:factool_visb} and ~\ref{fig:tot_visb} refer to the microservices without pipelining baseline.
The time blocks refer to continuous blocks of time when processing occurs on
this stage for any replica.
In FacTool, \sys generates search queries as claims are produced, and does
search per search query as queries are produced.
In Tree-of-Thought, \sys evaluates thoughts as they are proposed.
In contrast, the baselines have full barriers, waiting for all processing
within a single query to finish before proceeding to the next stage.

\subsubsection{Effects of Features in \sys}
\begin{figure}[t!]
     \centering
         \centering
         \includegraphics[width=\columnwidth]{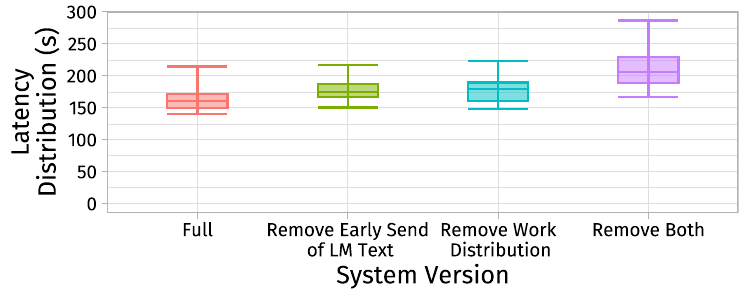}
         \caption{\sys, compared to versions of \sys with features removed, on
             the tree-of-thought workload; each box-whisker shows the
             distribution of sending 16 games, at rate 1 per minute, over two
             trials.
         Delayed language model sends alone and removing parallelization
         of work alone
     only increase median latency by about 13-18s, but removing them both
 increases median latency by 45s.}
         \label{fig:tot-ablations}
\end{figure}

This section explores two features in \sys: the ability to send output early
from a language model, and the ability to parallelize processing over
partial outputs across workers, in the Tree-of-Thought application.
Figure~\ref{fig:tot-ablations} shows the effect of taking each of these
features away, and both, from \sys.
Note that delaying sending output from the language model does not prevent the
system from executing on partial outputs in parallel; it just means the
language model stage delays sending all partial outputs until the generation
is over.
Removing parallelization refers to removing how \sys routes data in a \cc{pmap} call
(where data all has separate ancestry) to independent workers, as the ancestry
ID of the input data is different.
In Tree-of-Thought, this concretely means forcing all thought evaluations to
go through a single worker.
Removing either feature alone only increases median latency slightly (by
13-18s), but removing both features increases latency significantly by
45s.
When language model outputs are not sent early, work can still be parallelized
across workers to absorb the burst of work.
When language model outputs are streamed, but parallelization is not
available, the work can be absorbed over time.
Removing both slows down the application by creating short bursts of work.

\section{Related Work}

\para{Programming models and runtimes for compound AI systems}
Multiple groups have proposed both programming models and optimized runtimes for compound AI systems.
Programming models like LangChain, LangGraph, LlamaIndex, AutoGen and DSPy~\cite{ chase2022langchain, chase2022langgraph, liu2022llamaindex, AutoGen, khattab2023dspy} make it easy to build an AI system as a graph of composable modules, defined explicitly or inferred implicitly from Python code.
Many of them support explicit parallel calls and asynchronous execution.
However, none of them aim to optimize the performance of the application through techniques like streaming partial outputs and routing based on ancestry, as we do in \sys. \sys offers a similar graph-like, composable programming abstraction that automatically performs these optimizations.

In addition, many recent efforts have sought to optimize various components used in compound AI systems and some scenarios where the components interact.
LM serving engines like vLLM~\cite{kwon2023efficient}, Sarathi-Serve~\cite{sarathi-serve} and others focus on serving LM inference calls efficiently without awareness of how those calls may be composed in an application.
SGLang~\cite{zheng2023efficiently} goes further by providing a frontend API where the user can explicitly make batch calls with the same prefix and optimizes their submission to the serving engine, but does not consider streaming between multiple LM calls.
Ayo~\cite{ayo} considers streaming the output of language models into
retrievers, but does not provide a general-purpose abstraction for handling multiple granularities and tracking streaming dependencies across arbitrary components.
InferCept~\cite{InferCept} optimizes LM serving when tools are called during text generation to better reuse cached key-value embeddings for prefixes of a generation.
Conveyor~\cite{Conveyor} allows individual tools called by an LLM, such as a code interpreter, to take in streaming input (e.g., lines of code), but does not consider how to optimize and safely execute a graph of multiple LM and tool calls.
Murakkab~\cite{murakkab} and RAGServe~\cite{RAGServe} focus on optimizing the hardware choices, model choices, and parameter choices (e.g., number of documents retrieved) for a compound AI system to achieve quality-cost tradeoffs, but do not optimize streaming within the application.
Retrieval models and engines have also received significant performance optimization~\cite{hnsw,diskann,splade,santhanam2022plaid}.
\sys can leverage these components but uniquely focuses on optimizing data streaming within an application and contributes a programming model and an abstraction for tracking dependencies (nested ancestry) that enables this.

\para{Parallel data processing}
Systems including MapReduce~\cite{mr}, Dryad~\cite{dryad}, Ciel~\cite{ciel}, Cilk~\cite{cilk}, Legion~\cite{legion}, Apache Spark~\cite{spark} and parallel databases~\cite{dewitt-gray-parallel} all aim to execute a graph of operators on parallel resources.
Some of these can directly stream records between the operators~\cite{dewitt-gray-parallel,dryad}, while others persist intermediate data to stable storage~\cite{mr}.
However, these systems model data as collections of atomic records, which can be routed and tracked individually, instead of streams that may be consumed at different granularities, like the text outputs from an LM.
Thus, none of them consider the problem of deciding when to split data into partial outputs, routing them, and scheduling downstream execution to preserve semantics.

\para{Streaming systems}
Distributed streaming engines like Apache Flink~\cite{ApacheFlink},
Naiad~\cite{Naiad}, FlumeJava~\cite{flumejava} and Spark Streaming~\cite{dstreams} implement various strategies for parallel stream processing.
However, these systems focus on indefinitely long-running applications that need to pass data through a fixed graph of operators continuously, rather than short-lived requests that pass through dynamic computation graphs, like the ones in compound AI systems.
In addition, none of them explicitly models streams that can have partial outputs consumed at different granularities like the ones in LM applications.
Naiad~\cite{Naiad}'s logical timestamps are similar in spirit to \sys's nested
ancestry tags.
However, Naiad's timestamps are used to track a piece of data's
progress through a loop, while \sys's ancestry tags track data's hierarchy
(from nested pmap calls)
in the program structure.

\para{Other distributed computing frameworks}
A number of cluster computing frameworks aim to provide transparent access to distributed computation for non-expert users.
Ray~\cite{moritz2018ray}, Orleans~\cite{orleans} and Erlang~\cite{erlang} primarily target fine-grained, low-latency computation using an actor abstraction and explicit message passing.
PyWren~\cite{pywren} massively parallelizes work using serverless computation.
Like the parallel data processing frameworks we discussed, these systems do not explicitly model the outputs of an operator as a stream that can be consumed at different granularities.

\section{Conclusion}
This paper describes \sys, a framework that automatically executes
compound AI applications efficiently via pipelining and
parallelism, which matches or outperforms hand-optimized implementations of
four applications.
Compound AI applications contain a new data
processing pattern where language model components produce data streams that further components execute on in parallel.
To support streaming, \sys keeps tracks the provenance of data through the
dataflow, through a new
abstraction called nested ancestry.
\sys infers nested hierarchy  automatically, as it is coupled
with the programming model.
Nested ancestry is applicable to any data processing pipeline where components produce an unknown number of outputs, in a streaming fashion, and can help with
tracing, debugging, and executing these applications.

\bibliographystyle{ACM-Reference-Format}
\bibliography{cais-serving-paper.bib}

\end{document}